\documentclass[journal]{IEEEtran}
\usepackage{algorithmic}
\usepackage{algorithm}
\usepackage[utf8]{inputenc}
\usepackage[T1]{fontenc}
\usepackage{siunitx}
\usepackage[colorlinks=true, allcolors=blue]{hyperref}
\usepackage{amsmath}
\usepackage{amsfonts}
\usepackage{makecell}
\usepackage{soul} 
\usepackage{subcaption}
\usepackage{stfloats}
\usepackage{booktabs}

\usepackage{tikz,xcolor}
\definecolor{lime}{HTML}{A6CE39}
\DeclareRobustCommand{\orcidicon}
{
    \begin{tikzpicture}
    \draw[lime, fill=lime] (0,0) circle [radius=0.16] 
    node[white] {{\fontfamily{qag}\selectfont \tiny ID}};    \draw[white, fill=white] (-0.0625,0.095) circle [radius=0.007];    
    \end{tikzpicture}
    \hspace{0mm}}
\foreach \x in {A, ..., Z}{%
    \expandafter\xdef\csname orcid\x\endcsname{\noexpand\href{https://orcid.org/\csname orcidauthor\x\endcsname}{\noexpand\orcidicon}}
}

\newtheorem{myDef}{Definition}

\begin{document}

\title{PCG-based Static Underground Garage Scenario Generation}

\author{Wenjin Li, Kai Li

\thanks{Wenjin Li, Kai Li are with the Department of Computer Science and Technology, Southern University of Science and Technology, Shenzhen, 518055, China}
}

\markboth{Journal of \LaTeX\ Class Files,~Vol.
}%
{Shell \MakeLowercase{\textit{et al.}}: Bare Demo of IEEEtran.cls for IEEE Journals}

\maketitle

\begin{abstract}
Autonomous driving technology has five levels, from L0 to L5. 
Currently, only the L2 level (partial automation) can be achieved, and there is a long way to go before reaching the final level of L5 (full automation). 
The key to crossing these levels lies in training the autonomous driving model. 
However, relying solely on real-world road data to train the model is far from enough and consumes a great deal of resources. 
Although there are already examples of training autonomous driving models through simulators that simulate real-world scenarios, these scenarios require complete manual construction. 
Directly converting 3D scenes from road network formats will lack a large amount of detail and cannot be used as training sets. 
Underground parking garage static scenario simulation is regarded as a procedural content generation (PCG) problem. 
This paper will use the Sarsa algorithm to solve procedural content generation on underground garage structures.
\end{abstract}


%
\IEEEpeerreviewmaketitle

\section{Introduction}

According to a recent technical report by the National Highway Traffic Safety Administration (NHTSA), 94\% of road accidents are caused by human errors \cite{singh2015critical}. Against this backdrop, Automated Driving Systems (ADSs) are being developed with the promise of preventing accidents, reducing emissions, transporting the mobility-impaired, and reducing driving-related stress \cite{crayton2017autonomous}.

Autonomous driving simulation is an important part of ADSs. 
However, simulation lacks interactive and changeable scenarios \cite{lan2016development,lan2016developmentuav}. Researchers are still using authentic human-made ways to build one scenario for huge training.

Procedural Content Generation for Games (PCG-G) is the application of computers to generate game content, distinguish interesting instances among the ones generated, and select entertaining instances on behalf of the players \cite{Survey}. 
In our project, we consider the underground garage as the game content that should be generated. 
The problem can normally be divided into three parts. The first part is to create the digit grid map for each type of floor, as a PCG task. The second part is to convert each type of floor to the design diagram. 
The last part is to simulate the whole 3D scenario map depending on the design diagram. 
To simplify the simulation, we combine the last two parts as one part.

In reinforcement learning \cite{sutton1998barto}, an agent seeks an optimal control policy for a sequential decision-making problem. 
We regard the first part as a sequential decision-making problem. 
Markov decision processes (MDPs) are effective models for solving sequential decision-making problems \cite{boutilier1999decision} in uncertain environments. 
The agent's policy can be represented as a mapping from each state it may encounter to a probability distribution over the available actions \cite{lan2016action}.  
Generalized policy iteration (GPI) was demonstrated as a class of iterative algorithms for solving MDPs in \cite{sutton1998barto}. 
It contains policy iteration (PI) and value iteration (VI) as special cases and has both advantages of PI and VI. 
Temperal-difference \cite{TD1} is the specific implementation of GPI \cite{Book}. 
TD methods are guaranteed to converge in the limit to the optimal action-value function, from which an optimal policy can be easily derived. 
A classic TD method is Sarsa \cite{sarsa}. 
The on-policy algorithm, in which policy evaluation and policy improvement are identical, has important advantages. 
In particular, it has stronger convergence guarantees when combined with function approximation, since off-policy approaches can diverge in that case. 
In this paper, we use the Sarsa algorithm to create a digit grid map.

Simulation is an important step during the conversion \cite{lan2019evolutionary}. 
We consider the simulator can generate test scenarios automatically, including static buildings, dynamic traffic flow, and real-time calculated lighting and weather. 
This paper aims to solve the static scene generation problem.

\section{Related Work}
\label{sec:related_work}
Abdullah \cite{Abdullah} compared the space utilization efficiency of diagonal, parallel, and perpendicular parking methods and concluded that perpendicular parking methods have the highest number of spaces, using a university as a specific example.
Sawangchote \cite{Sawangchote} developed a heuristic algorithm for the layout of parking spaces in small-scale garages based on the space utilization of different parking methods;
Xu Hanzhe \cite{Xu} carries out a parking space layout design based on a greedy algorithm to study the influence of irregular contours and obstacles on the layout of parking spaces and get the layout plan with the most number of parking spaces.
Julian Togelius \cite{composition} finds that the result of the composition of a bunch of different algorithms is better than the result of any single algorithm and He used answer set programming to do procedure content generation. 
Huimin Wang \cite{RLpp} has previously proposed a model-based reinforcement learning algorithm to implement the path planning problem. The path planning problem has similar features when it applies to the specialized PCG problem. We consider that generation on a garage can use the method of path planning on agent moving. Besides, Arunpreet Sandhu \cite{WFC} comes up with the WFC algorithm to generate similar images.
Akatsu \cite{Akatsu} provides an idea for evaluating underground garage structures by feeding a series of indicators obtained from a realistic traffic survey into a modeled underground garage structure to obtain a series of evaluation results.

\section{Methodology}
\label{sec:methodology}
\subsection{Overall}
We consider dividing the underground garage construction into two main parts, PCG task and simulation. Notations using throughout this report are as follows:\par
\begin{table}[!ht]
\begin{center}
\caption{The symbol table.}
\begin{tabular}{c|c}
\textbf{symbol} & \textbf{definition}\\ \hline
\(\mathcal{S}(i, j) \) & \makecell{Structure matrix, each element of which represents the\\ type of objects placed in the corresponding area} \\
\(R(i, j) \) & \makecell{A matrix of row values, each element of which represents\\ the horizontal width of the corresponding area} \\
\(C(i, j) \)&  \makecell{A matrix of column values, each element of which represents\\ the vertical width of the corresponding area} \\
\(h \) & Height of S(i,j) \\
\(w \) & Width of S(i,j) \\
\(M(i,j) \) & Manhattan distance between i and j \\
\(\epsilon \) & Squares with entrance or exit \\
\(o \) & Squares with obstacle \\
\(\xi \) & Frontier square \\
\(\eta \) & Inner square \\
\(\rho_{i} \) & Eight Squares around square I \\
\(\theta_{i} \) & Four Squares around square I \\
\(\zeta \) & Squares in underground garage floor plan \\
\(\psi \) & Squares not in underground garage floor plan \\
\(\delta \) & One solution in the solution space \\
\( F(\delta) \) & The fitness value of an individual \\
\(\sigma_{i} \) & \makecell{The minimum Manhattan distance between two squares} \\
\(N_S \) & The number of parking spots of structure matrix S \\
\(T_S \) & The average parking time of structure matrix S \\
\(U_S \) & The numebr of unused squares of structure matrix S \\
\(y^r \) & The real value of a structure matrix \\
\(y^{'} \) & The estimate value of a structure matrix \\
\(k_i\) & Coefficients of evaluation function indicators \\ \bottomrule
\end{tabular}
\end{center}
\end{table}
Since the most important thing in static underground garage scenario generation problems is the planning of parking stalls. For parking space planning problem, it is essentially an optimization problem of object placement, the objects to be placed will have the following distinction: 
\begin{itemize}
    \item static object: object's position will not change after confirming the position
    \item dynamic object: objects can wait for further optimization after confirming the position of static objects
\end{itemize}

Now we only need to consider the dynamic object distribution, in order to better describe the entire underground garage object planning situation, here we rasterize the underground garage by using three matrices $S_{i,j}, R_{i,j}, C_{i,j}$ to describe the state of an underground garage. 
In this paper, we will use reinforcement learning to plan the distribution of dynamic objects, by combining the distribution with the distribution of static objects to obtain the $S_{i,j}$ as the result of parking space planning, and finally combine the $R_{i,j}$ and $C_{i,j}$ as the plane structure of the static underground garage to pass into the Unity3D engine for 3D modeling to finally generate the static underground garage scenario.

We provide the following requirements for a reliable garage:
\begin{itemize}
    \item Reality: The generated basement structure needs to adapt to real-world standards (such as national standards and regulations)
    \item Feasibility: Ensure that at least one route to any exit and entrance can be found for each parking space arranged in the basement structure
    \item Randomness: The structure and contour of the basement are randomly generated, and the solution generated each time will change according to the change of the random process
    \item Bijection: Each generated basement structure has a unique corresponding random process, and this random process must correspond to a unique basement structure
    \item Customizability: The structure of the basement can be self-defined
\end{itemize}

\subsection{Static objects generation}
 First, we give a definition of structure matrix $\mathcal{S}(i,j)$:\par
\begin{equation}
\label{deqn_ex1}
\mathcal{S}(i,j)=\left\{
\begin{aligned}
0&, & & \text{parking space or free space} \\
-1&, & & \text{obstacle} \\
1&, & & \text{lane} \\
2&, & & \text{entrance} \\
3&, & & \text{exit} \\
\end{aligned}
\right.
\end{equation}
 At the beginning of getting this matrix, we should confirm the location of those static objects, which can be divided into three steps: contour generation, entrance and exit generation, and obstacle generation.
 
First, we need to generate the contour of the underground garage. Divide a $w\times h$ rectangle into $w\times h$ blocks and each block has a width and height of 1. We consider generate n groups of 2n points in this rectangle and use the line of two points of each group as the diagonal of the rectangle to generate a rectangle and then after expand all rectangles to its corresponding squares, We will treat the concatenation of all rectangles as a generated underground garage contour. The following algorithm shows the generation of underground garage contour.
\begin{algorithm} 
	\caption{contour generation} 
	\label{alg3} 
	\begin{algorithmic}
	\WHILE{$recs$ are not overlap}
		\STATE $recs \gets GetRectangles()$ 
		\FOR{$i$ in $recs$}
		    \FOR{$j = i+1$ in $recs$}
		        \IF{$recs[i],recs[j]$ not $Overlap()$}
		            \STATE break
		        \ENDIF
		    \ENDFOR
		\ENDFOR
	\ENDWHILE
	\FOR{$i$ in recs}
	   \STATE $plane = plane \cup Reshape(recs[i])$
	\ENDFOR
	\RETURN  $plane$
	\end{algorithmic} 
\end{algorithm}

After contour generation, we can get all squares in the floor plan, which mean we get $\zeta$ and $\psi$ and then assign values to all those squares in $\zeta$ and $\psi$:
\begin{equation}
\label{deqn_ex1}
\mathcal{S}(\zeta) = 0
\end{equation}
\begin{equation}
\label{deqn_ex1}
\mathcal{S}(\psi) = -1
\end{equation}
\par
Secondly, we need to determine the position of the entrance and exit. After contour generation, in order to generate a reliable position of entrance and exit, we give a definition of $\xi$ and $\eta$. A frontier square needs to satisfy the following conditions:
\begin{equation}
\label{deqn_ex1}
\mathcal{S}(\xi) = 0
\end{equation}
\begin{equation}
\label{deqn_ex1}
\sum_{i=1}^{8} \mathcal{S}(\rho_{\xi}) < 0
\end{equation}
An inner square needs to satisfy the following conditions:\par
\begin{equation}
\label{deqn_ex1}
\mathcal{S}(\eta) = 0
\end{equation}
\begin{equation}
\label{deqn_ex1}
\sum_{i=1}^{8} \mathcal{S}(\rho_{\eta}) = 0
\end{equation}
\\
Since entrances and exits can only be generated in $\xi$ and cannot be generated on the corners of $\xi$, in this condition, we only generate entrance and exit on those squares satisfy the following condition:
\begin{equation}
\label{deqn_ex1}
\epsilon \in \xi
\end{equation}
\begin{equation}
\label{deqn_ex1}
\sum_{i=1}^{8} \mathcal{S}(\rho_{\epsilon}) = -3
\end{equation}
\par
\begin{equation}
\label{deqn_ex1}
M(\epsilon_{i},\epsilon_{j}) \geq \sigma_{1}
\end{equation}
\par
Thirdly, we need to consider the position of obstacles in this underground garage. We only generate obstacles on those squares satisfying the following conditions:
\begin{equation}
\label{deqn_ex1}
o \in \eta
\end{equation}
\begin{equation}
\label{deqn_ex1}
M(o_{i},o_{j}) \geq \sigma_{2}
\end{equation}
\par

\subsection{Reinforcement Learning}

Reinforcement learning (RL) is a basis to solve our PCG problem. In this paper, we first focus on finite Markov decision processes (finite MDPs).

\begin{myDef}
A finite Markov decision process can be
represented as a 4-tuple $M = \{S, A, P, R\}$, where $S$ is a
finite set of states; $A$ is a finite set of actions; $P : S\times R \times S \times A \to [0, 1]$ is the probability transition function; and $R : S \times A \to \mathcal{R}$ is the reward function. In this paper, we denote the probability of the transition from state $s$ to another state $s'$ when taking action a by $P(s', r|s, a)$ and the immediate reward received after the transition by $r_s^a$ \cite{ferns2003metrics}. 
\end{myDef}

A policy is defined as a mapping, $\pi: S\times A\to [0,1]$. In this paper, we use $\pi(s)$ to represent the action $a$ in state $s$ under the policy $\pi$. To measure the quality of a policy, action-value function, $q_\pi(s, a)$ is used to estimate the expected long-term cumulative reward of taking action $a$ in state $s$ under a policy $\pi$. It is formally defined as:

\begin{equation}
    q_\pi(s,a)=\mathbb{E}_\pi[\sum_{k=0}^\infty\gamma^kR_{t+k+1}| S_t=s, A_t=a]
\end{equation}

where $\gamma$ is a discount factor, $R_t$ is the reward at time-step $t$, and $\Bbb{E}_\pi$ is the expectation with respect to the policy $\pi$.

The goal is to find an optimal policy $\pi_*$ which maximizes the expectation of long-time discounted cumulative reward from any starting state $s\in S$:

\begin{equation}
    \pi_*=\operatorname*{argmax}_{\pi} \Bbb{E}_\pi [\sum_{t=0}^\infty \gamma^t R_t|s_0=s]
\end{equation}

In this paper, we format PCG as an optimization problem \cite{lan2022time,lan2021learning,lan2021learning}, which is represented as a 2-tuple $(M, E )$, where $M$ is finite MDPs which can generate one 2D integer array and $E$ is an evaluation function which evaluates the quality of array. We have one agent with policy $\pi$. It will tack action in state $s$ and send a message to the environment. 
The environment receives the message and changes the state to the next state and sends rewards to the agent.

Finally, the agent and environment produce a finite Markov decision array:
\begin{equation}
    S_0,A_0, R_1, S_1, A_1, R_2, S_2, A_2, R_3,\dots, S_{T-1}, A_{T-1}, R_T
\end{equation}
where $T$ is the termination time. Evaluation function $E$ is calculated from $M$
\begin{equation}
    E=\sum_{t=1}^{T-1} R_t
\end{equation}
$R_T$ is always a negative value and it is not included in $E$. In other words, we come back to the previous unfailed state to compute $E$.

Generalized policy iteration (GPI) contains two processes, policy evaluation (E) and policy improvement (I):
\begin{equation}
    \pi_0\stackrel{\text{E}}{\to} q_{\pi_0}\stackrel{\text{I}}{\to} \pi_1\stackrel{\text{E}}{\to} q_{\pi_1}\stackrel{\text{I}}{\to} \pi_2\stackrel{\text{E}}{\to}\dots\stackrel{\text{I}}{\to} \pi_{*}\stackrel{\text{E}}{\to} q_{*}
\end{equation}
where $q_{\pi_i}$ is action value function under $\pi$ at episode $i$. The process is terminated when $q$ and $\pi$ converges to $q_*$ and $\pi_*$. For Sarsa algorithm, policy evaluation and policy improvement are carried out simultaneously in each episode.

The agent and environment in MDP are clear. Our design is divided into two sections. 
In the first section, we design the MDP for our PCG task. In the other section, we design the environment penalty based on the principle of parking lot design.

\subsection{Sarsa}
We use the Sarasa algorithm to solve the PCG task. First, we define the parameters of MDPs. We consider a car in a 2D place as an agent to perform a colouring task, which colours the undefined square to a lane spuare. Agent's state at timestamp $t$ is defined as the multiple dimensional vectors:
\begin{equation}
    S_t=(D, M, A_{t-1})
\end{equation}
Where $D$ is a 4-dimensional vector that each element point to the distance between the free space, border, or obstacle and agent in the direction, $M$ is a 25-dimensional vector that symbols to the perception range of the agent. It satisfies that all points have a Manhattan distance of less than 2 from the agent.

The agent takes action from the action set
\begin{equation}
    A=\{\textbf{UP}, \textbf{DOWN}, \textbf{LEFT}, \textbf{RIGHT}, \textbf{STAY}\}
\end{equation}

The goal is to colour the road as much as possible until it comes back to the start and takes action $\textbf{STAY}$, leading to a terminate state. Agent receives rewards depending on the increment of the number of parking spaces. The agent also receives a penalty for some wrong actions.
\begin{figure}  \centering
    \begin{minipage}{0.7\linewidth}  \centering
      \includegraphics[width=\linewidth]{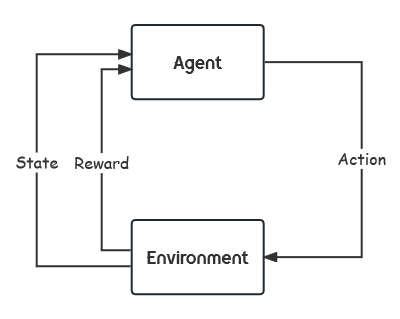}
      \caption{MDPs framework}
      \label{fig:state}
    \end{minipage}
\end{figure}
To evaluate one policy $\pi$, we predict one Markov decision array containing $S, A, R$ for each episode. We update $q(S_t, A_t)$ during the prediction, following the function:
\begin{equation} \footnotesize
    q(S_t, A_t) = q(S_t, A_t) + \alpha \times (R_{t+1} + \gamma \times q(S_{t+1}, A_{t+1})-q(S_t, A_t))
\end{equation}
where $\alpha$ and $\gamma$ are parameters, with $0\le \alpha, \gamma\le 1$.

We use greedy method to improve one policy:
\begin{equation}
    \pi(s)=\operatorname*{argmax}_{a} q(s,a)
\end{equation}
where $\pi(s)$ is the greedy action under policy $\pi$. We consider using $\epsilon$-greedy to take action, where the agent has $\epsilon$ chance of taking greedy action with maximum value otherwise taking action equivalently. The probability of taking greedy action $\pi(s)$ in state $s$ is:
\begin{equation}
    p(s, \pi(s)) = (1-\epsilon)+\frac{\epsilon}{|A|}
\end{equation}

\subsection{Penalty design}
The principle of parking lot design has been proposed for optimizing parking area space.
\begin{itemize}
    \item Use rectangular areas where possible
    \item Make the long sides of the parking areas parallel
    \item Design so that parking stalls are located along the lot's perimeter
    \item Use traffic lanes that serve two rows of stalls
\end{itemize}

\begin{figure*} [!ht]  \centering
    \begin{minipage}{0.47\linewidth}       \centering
      \includegraphics[width=\linewidth]{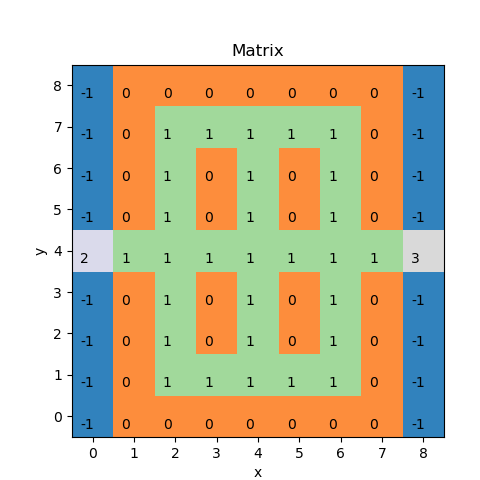}
      \caption{Conforming Map}
      \label{fig:conform1}
    \end{minipage}
    \begin{minipage}{0.47\linewidth}       \centering
      \includegraphics[width=\linewidth]{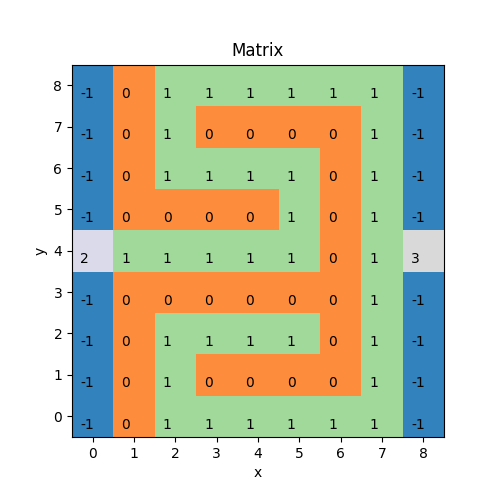}
      \caption{Not conforming Map}
      \label{fig:nconform1}
    \end{minipage}
    \hspace{0.05\linewidth}
\end{figure*}

\autoref{fig:conform1} conforms the above principle, where green square refers to lane square, orange square refers to parking square or free square, and white square refers to entrance or exit. Contrary to \autoref{fig:conform1}, \autoref{fig:nconform1} has many problems: no cycle, existing non-rectangular and non-parallel areas, and many lanes serving only one row of the stall.

The agent can not only receive a reward after the action but also a certain penalty we defined. The reasonable penalty guides agents to do actions they want. Based on the design principle, we propose several penalties below:
\begin{itemize}
    \item Turn-back penalty when the agent takes the opposite action from the last action.
    \item Interval penalty based on the interval of the same actions.
    \item Wheeling penalty at an improper position with a certain direction. 
    \item Step penalty for each timestamp to prevent agents from cycling consistently.
\end{itemize}

\subsection{Convert matrix to simulated underground garage}
After generating structure matrix $\mathcal{S}(i,j)$, we need to convert this matrix to a simulated underground garage. Here we first atomize the elements of the matrix, we define the below equation:
\begin{equation}
\label{deqn_ex1}
n = \sum_{i=1}^{4} \mathcal{S}(\theta_{\eta})
\end{equation}
and for any square $\eta$, if:
\begin{equation}
\label{deqn_ex1}
\mathcal{S}(\eta) = 1
\end{equation}
we define $\eta$ as:
\begin{equation}
\label{deqn_ex1}
\eta=\left\{
\begin{aligned}
Crossroads&, & & \text{n = 4} \\
T-Junctions&, & & \text{n = 3} \\
Straight road&, & & \text{n $\leq$ 2} \\
\end{aligned}
\right.
\end{equation}
and if:
\begin{equation}
\label{deqn_ex1}
\mathcal{S}(\eta) = 0
\end{equation}
we define $\eta$ as different types in Figure 2:
\begin{equation}
\label{deqn_ex1}
\eta=\left\{
\begin{aligned}
Type1&, & & \text{n $\geq$ 3 or across n = 2} \\
Type2&, & & \text{adjacent n = 2,} \\
Type3&, & & \text{n = 1} \\
Type4&, & & \text{n = 0} \\
\end{aligned}
\right.
\end{equation}
\begin{figure*}   \centering
    \includegraphics[width=0.95\textwidth]{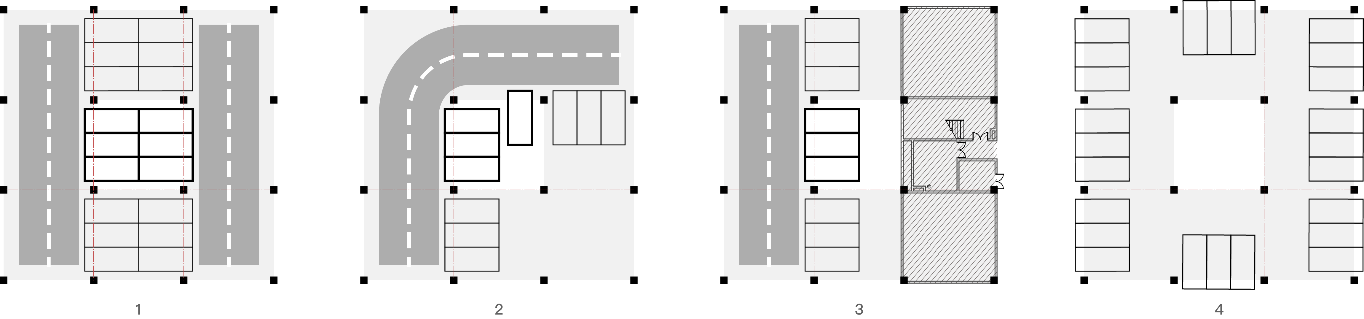}
    \caption{Parking Space Type}
    \label{fig:PST}
\end{figure*}

Then, we only need to model each type of square $\eta$ in the simulator and use scripts to construct the simulated underground garage.

\subsection{Construction of underground garage structure}
We know that autonomous vehicles typically use multiple types of sensors to collect and process environmental information to support the vehicle's decision-making and control systems \cite{xu2019online}. 

The parking garage structure we generate is intended to provide training scenarios for autonomous vehicles, and the information collected during autonomous vehicle training comes from the simulated scenes, such as the lighting of light sources, the materials of various object surfaces, and information on the different light reflections of objects in the scene, and so on \cite{lan2019simulated}. If we can better simulate the various objects in these scenes, the amount of information contained in the overall static parking garage scene will be greater, and it will better provide training data for autonomous vehicles, achieving better training effects.

The construction details of a static underground parking garage mainly include object surface texture mapping, such as:
\begin{enumerate}
    \item Lane marking texture mapping
    \item Wall texture mapping
    \item Floor texture mapping
    \item Lighting texture mapping
\end{enumerate}
\par

As well as collision bodies in the underground parking garage, such as:
\begin{enumerate}
    \item Column mesh collision body
    \item Speed bump collision body
    \item Parking barrier
\end{enumerate}
\par
And here we give the detailed procedure of underground garage generation in Unity3D:
\begin{enumerate}
    \item The structure matrix $\mathcal{S}_(i,j)$ previously generated by using reinforcement learning is used as the generated underground structure, and the $R_(i,j)$ and $C_(i,j)$, which define the length and width of each plot of land in reality, are passed as input into Unity3D engine.
    \item In the Unity engine, each different state of the land is first modeled, and then the entire underground plane is automatically generated based on the arrangement of elements in the specific structure matrix.
    \item After generating the plane, three-dimensional information such as walls, pillars, ceilings, obstacles, etc. are further generated based on the outline of the underground structure.
    \item According to the generated structure, more detailed descriptions are made, such as light tubes, ventilation ducts, and other underground details.
    \item According to the demand, some objects that may appear underground, such as parked vehicles and no parking signs, are randomly generated.
\end{enumerate}

\section{Experimental setup}
\label{sec:setup}
\subsection{Evaluation}
After generating the underground garage structure, we need to evaluate it, but there is no unified and credible standard for the evaluation function. So we proposed the following three dimensions to describe the value of the underground garage structure by combining the evaluation system of several papers:
\begin{itemize}
    \item the number of the parking spot
    \item the average parking time
    \item the number of unused squares
\end{itemize}
\par
So the evaluation function is like:
\begin{equation}
\label{deqn_ex1}
y^{'} = k_1 * N_S + k_2 * T_S + k_3 * U_S
\end{equation}
\par
To obtain the proportion of weights accounted for by each of these three criteria, here we assume that there exists a corresponding evaluation function for a certain underground garage structure, and the value distribution of all solutions for that structure is roughly Gaussian distributed. 
\par
Based on this, we can know that if we have enough sampling points and judge the value size relationship of the structure in the sampling points, we can correspond these sampling points to the Gaussian distribution curve one by one, and then make the estimated value order of the sampling points the same as before by adjusting the weights of our evaluation function, so that we get an evaluation function with a certain degree of confidence, and when more and more points are sampled, the final evaluation function will be more credible.
\par
Here, we sampled a series of more representative experimental results and derived the above values for the three coefficients:
\begin{equation}
\label{deqn_ex1}
y^{'} =  N_S + (-5) * T_S + (-1) * U_S
\end{equation}
\par
We conducted a 5000-episode cycle test for Sarsa algorithm with one garage contour. For each episode, we save the matrix and evaluation on it to the dictionary. In the end, we select top 200 matrix with high evaluation function value.

\subsection{Simulation of Underground Garage}
The main hardware devices used in the simulation to generate the underground garage scenario are: CPU: Intel(R) Core(TM) i7-10750H CPU @ 2.60GHz, GPU: NVIDIA GeForce GTX 1650 and the software are: Unity3D 2021.3.5f1c1, Visual Studio 2022

\section{Results}
\label{sec:results}
\subsection{Sarsa Result}
\begin{figure*} [!ht]\centering
    \begin{minipage}{0.33\linewidth}  \centering
      \includegraphics[width=\linewidth]{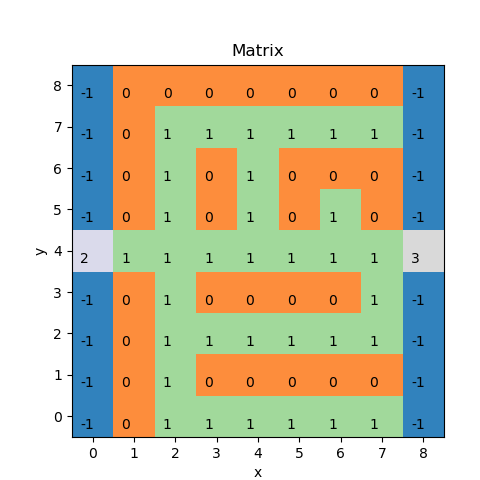}
      \caption{rank 1}
      \label{fig:rank1}
    \end{minipage} \hspace{-0.5cm}
    \begin{minipage}{0.33\linewidth} \centering
      \includegraphics[width=\linewidth]{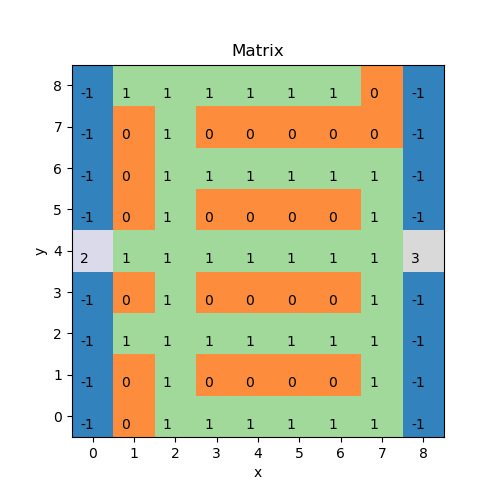}
      \caption{rank 2}
      \label{fig:rank2}
    \end{minipage} \hspace{-0.5cm}
    \begin{minipage}{0.33\linewidth} \centering
      \includegraphics[width=\linewidth]{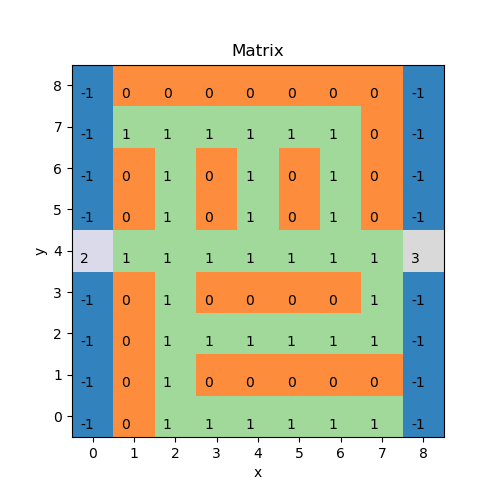}
      \caption{rank 3}
      \label{fig:rank3}
    \end{minipage}
    \begin{minipage}{0.33\linewidth} \centering
      \includegraphics[width=\linewidth]{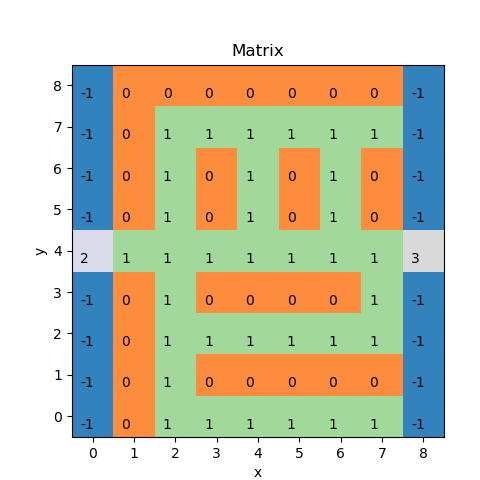}
      \caption{rank 4}
      \label{fig:rank4} 
    \end{minipage} \hspace{-0.5cm}
    \begin{minipage}{0.33\linewidth} \centering
      \includegraphics[width=\linewidth]{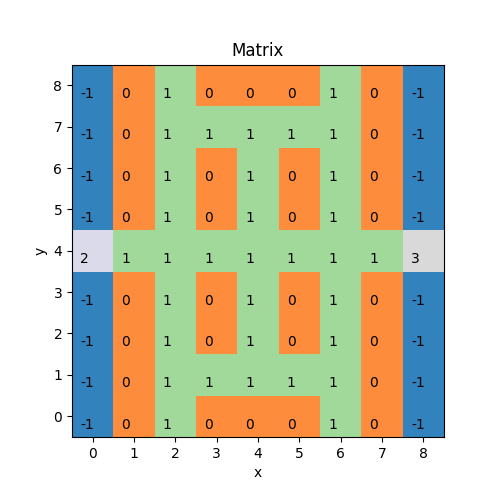}
      \caption{manual 1}
      \label{fig:manual1}
    \end{minipage}
    \begin{minipage}{0.33\linewidth} \centering
      \includegraphics[width=\linewidth]{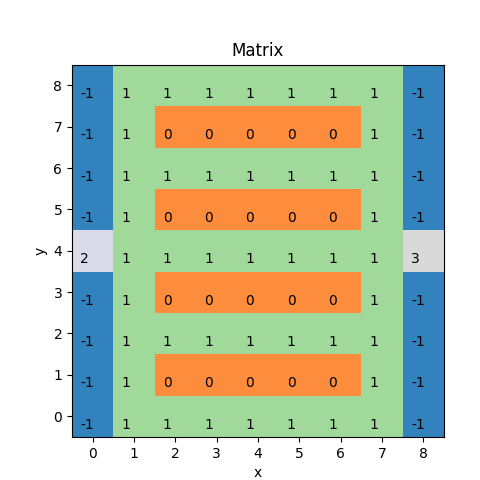}
      \caption{manual 2}
      \label{fig:manual2}
    \end{minipage}
    \begin{minipage}{0.95\linewidth} \centering
      \includegraphics[width=\linewidth]{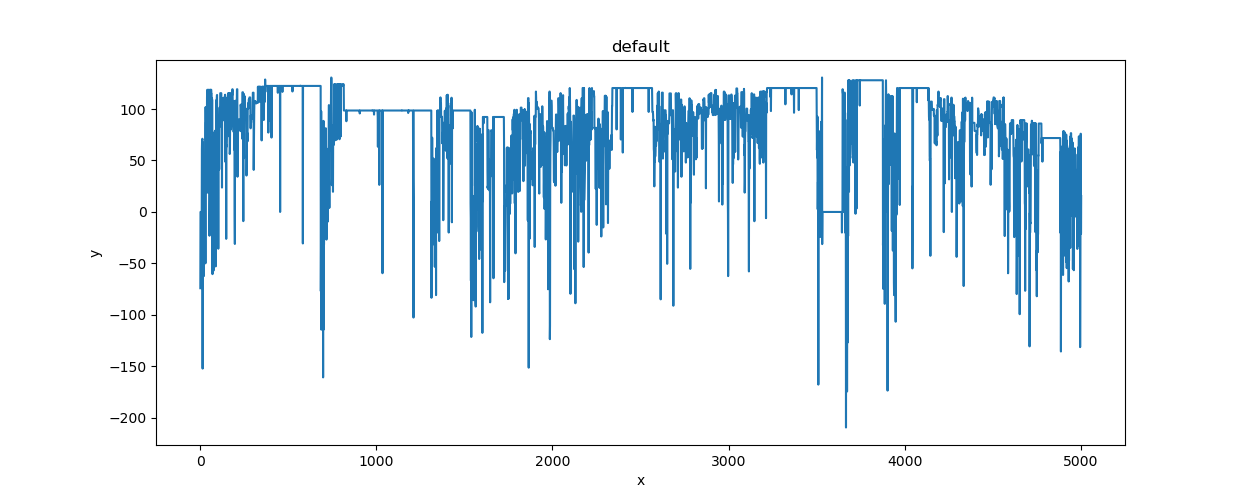}
      \caption{$G$-epi graph}
      \label{fig:repay}
    \end{minipage}
\end{figure*}

\autoref{fig:repay} indicate that the agent easily achieves the local limit at episode 400. Then it straight down to a small value. It maintains a trend of first converging to the limit and then sharply decreasing. It will keep searching for a solution if the test doesn't stop.

However, we observed that as the number of episodes increases, there are instances where the agent obtains lower payoffs. This can be attributed to the $\epsilon$-greedy strategy, which sometimes leads the agent directly to the termination state. To increase the converge rate, We make the $\epsilon$ decrease slowly. We also refresh the value of $\epsilon$ if the matrix keeps at 100 consecutive episodes.

\autoref{fig:rank1} shows the matrix with the highest evaluation value during the test. It is slightly inferior to \autoref{fig:manual1} and \autoref{fig:manual2} manually constructed.

\begin{table}[htbp] \centering 
	\caption{Evaluation Table} 
	\label{table2} 
	\begin{tabular}{|c|c|c|c|c|c|c|}  \hline  
		& & & & & & \\[-6pt] 
		Matrix id&rank1&rank2&rank3&rank4&manual1&manual2 \\  \hline
		& & & & & & \\[-6pt]
		$y'$&62.18&61.71&61.62&61.62&61.18&54.04 \\		\hline
	\end{tabular}
\end{table}

\subsection{Simulated Underground Garage}
\begin{figure}[!htp] \centering
    \includegraphics[width=\linewidth,trim={10 10 10 30},clip]{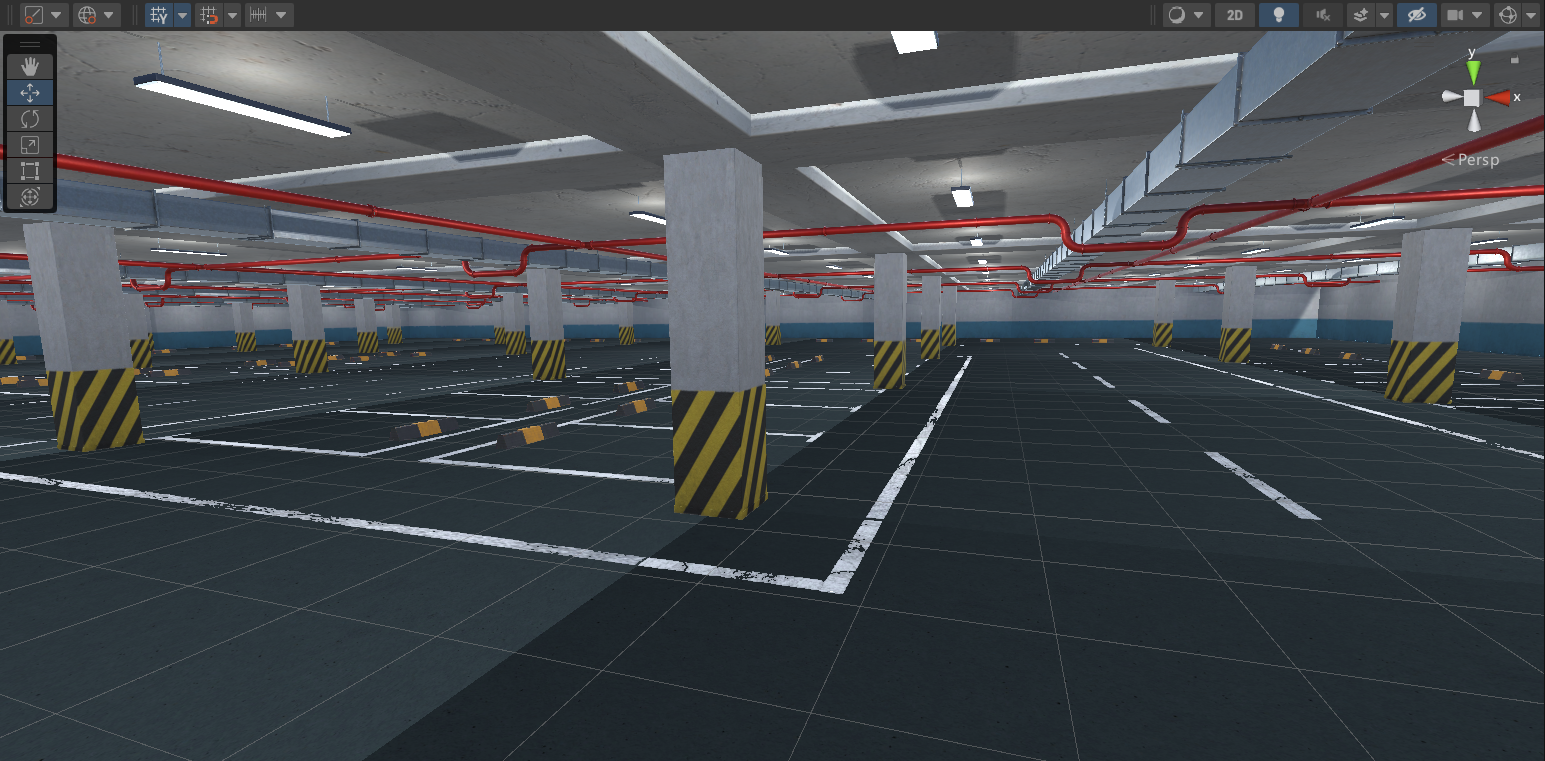}
    \caption{
    \centering{3D underground garage simulation}}
    \label{fig4}
\end{figure}

\begin{figure}[!ht]   \centering
    \includegraphics[width=\linewidth,trim={50 30 50 30},clip]{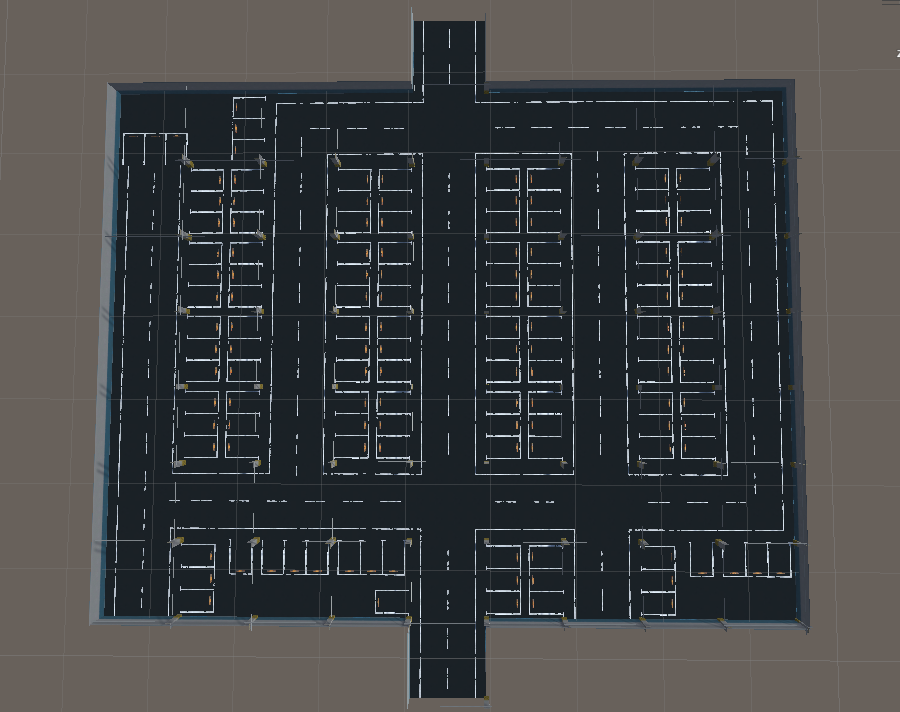}
    \caption{
    \centering{Simulation of Fig. 6}  }
    \label{fig5}
\end{figure}

\autoref{fig4} shows the underground garage model simulated by modelling the structure matrix generated by the above reinforcement learning algorithm for 3000 iterations as input.
\autoref{fig5} shows the underground garage model simulated by modelling the structure matrix generated by the above reinforcement learning algorithm for 3000 iterations as input.

\section{Discussion}
\label{sec:discussion}
For the evaluation function, there is no unified credible evaluation function, and the coefficient given in this paper is only a fitting operation for the real value curve. At the same time, since the structure of an underground garage with different contours has an impact on the three evaluation indexes we selected, the value of the coefficients for different contours may also be inconsistent, which may require more sampling and training through neural networks to come up with the coefficients for each underground garage contour later \cite{lan2022vision,sun2022multi,sun2023marine}.

However, happily, we were able to correctly evaluate the generated underground garage parking space structure according to the evaluation function obtained from the sampling on the 7*9 square contour, as it can be seen that Fig. 5 and Fig. 6 are the manually designed structures considered to be of higher value according to the cognitive design, and Fig. 1 to Fig. 4 are the top four structures of value filtered according to the evaluation function from the results of the algorithm generating 5000 episodes, and it can be seen that the filtered structures, although not perfect, can meet several of the most basic requirements in designing an underground garage parking space, and are indeed a little more valuable than the manually designed structures.

\section{Conclusions}
\label{sec:conclusion}
Sarsa, an on-policy TD algorithm, performs well in this paper. It can generate reliable graphs eventually. However, the state set is so large that it can not converge into one solution that reaches the highest repayment.

This study demonstrates the feasibility of using reinforcement learning to programmatically generate underground garage grid maps. We have yet to reach a target that can generate a reliable underground garage based on some contour. PCG of underground garage design has a long way to go.

In terms of simulation, we are currently able to construct the corresponding 3D underground parking garage and the generated garage has certain details: real-time lighting, ventilation ducts, column network structure, etc.. The current garage details such as various pipe layouts are not yet practical and various scene elements can be further rendered to achieve a more realistic effect. This will allow us to further enhance the accuracy and reliability of the generated underground garage maps. These findings provide valuable insights for the development of intelligent underground garage planning and design tools.
In the future, we will extend this work with other AI technologies, such as classification \cite{lan2022class,gao2021neat}, knowledge graphs \cite{liu2022towards,lan2022semantic}, deep learning \cite{lan2019evolving,lan2018real}.










\bibliographystyle{IEEEtran}
\bibliography{bibliography}

\begin{thebibliography}{10}
\providecommand{\url}[1]{#1}
\csname url@samestyle\endcsname
\providecommand{\newblock}{\relax}
\providecommand{\bibinfo}[2]{#2}
\providecommand{\BIBentrySTDinterwordspacing}{\spaceskip=0pt\relax}
\providecommand{\BIBentryALTinterwordstretchfactor}{4}
\providecommand{\BIBentryALTinterwordspacing}{\spaceskip=\fontdimen2\font plus
\BIBentryALTinterwordstretchfactor\fontdimen3\font minus
  \fontdimen4\font\relax}
\providecommand{\BIBforeignlanguage}[2]{{%
\expandafter\ifx\csname l@#1\endcsname\relax
\typeout{** WARNING: IEEEtran.bst: No hyphenation pattern has been}%
\typeout{** loaded for the language `#1'. Using the pattern for}%
\typeout{** the default language instead.}%
\else
\language=\csname l@#1\endcsname
\fi
#2}}
\providecommand{\BIBdecl}{\relax}
\BIBdecl

\bibitem{singh2015critical}
S.~Singh, ``Critical reasons for crashes investigated in the national motor
  vehicle crash causation survey,'' Tech. Rep., 2015.

\bibitem{crayton2017autonomous}
T.~J. Crayton and B.~M. Meier, ``Autonomous vehicles: Developing a public
  health research agenda to frame the future of transportation policy,''
  \emph{Journal of Transport \& Health}, vol.~6, pp. 245--252, 2017.

\bibitem{lan2016development}
G.~Lan, Z.~Luo, and Q.~Hao, ``Development of a virtual reality teleconference
  system using distributed depth sensors,'' in \emph{2016 2nd IEEE
  International Conference on Computer and Communications (ICCC)}.\hskip 1em
  plus 0.5em minus 0.4em\relax IEEE, 2016, pp. 975--978.

\bibitem{lan2016developmentuav}
G.~Lan, J.~Sun, C.~Li, Z.~Ou, Z.~Luo, J.~Liang, and Q.~Hao, ``Development of
  uav based virtual reality systems,'' in \emph{2016 IEEE International
  Conference on Multisensor Fusion and Integration for Intelligent Systems
  (MFI)}.\hskip 1em plus 0.5em minus 0.4em\relax IEEE, 2016, pp. 481--486.

\bibitem{Survey}
M.~Hendrikx, S.~Meijer, J.~Van Der~Velden, and A.~Iosup, ``Procedural content
  generation for games: A survey,'' \emph{ACM Trans. Multimedia Comput. Commun.
  Appl.}, vol.~9, no.~1, feb 2013.

\bibitem{sutton1998barto}
R.~Sutton, ``Barto:“reinforcement learning: An introduction”,'' \emph{IEEE
  Trans. Neural Netw}, vol.~9, p. 1054, 1998.

\bibitem{boutilier1999decision}
C.~Boutilier, T.~Dean, and S.~Hanks, ``Decision-theoretic planning: Structural
  assumptions and computational leverage,'' \emph{Journal of Artificial
  Intelligence Research}, vol.~11, pp. 1--94, 1999.

\bibitem{lan2016action}
G.~Lan, Y.~Bu, J.~Liang, and Q.~Hao, ``Action synchronization between human and
  uav robotic arms for remote operation,'' in \emph{2016 IEEE International
  Conference on Mechatronics and Automation}.\hskip 1em plus 0.5em minus
  0.4em\relax IEEE, 2016, pp. 2477--2481.

\bibitem{TD1}
R.~S. Sutton, ``Learning to predict by the methods of temporal differences,''
  \emph{Machine learning}, vol.~3, pp. 9--44, 1988.

\bibitem{Book}
R.~S. Sutton and A.~G. Barto, \emph{Reinforcement Learning: An Introduction},
  1988.

\bibitem{sarsa}
H.~van Seijen, H.~van Hasselt, S.~Whiteson, and M.~Wiering, ``A theoretical and
  empirical analysis of expected sarsa,'' in \emph{2009 IEEE Symposium on
  Adaptive Dynamic Programming and Reinforcement Learning}, 2009, pp. 177--184.

\bibitem{lan2019evolutionary}
G.~Lan, J.~Chen, and A.~Eiben, ``Evolutionary predator-prey robot systems: From
  simulation to real world,'' in \emph{Proceedings of the genetic and
  evolutionary computation conference companion}, 2019, pp. 123--124.

\bibitem{Abdullah}
K.~Abdullah, N.~H. Kamis, N.~F.~N. Azahar, S.~F. Shariff, and Z.~C. Musa,
  ``Optimization of the parking spaces: A case study of dataran mawar, uitm
  shah alam,'' in \emph{2012 IEEE Colloquium on Humanities, Science and
  Engineering (CHUSER)}, 2012, pp. 684--687.

\bibitem{Sawangchote}
P.~Sawangchote and T.~Yooyativong, ``Automated parking area optimization for
  garage construction using geometric algorithm,'' in \emph{2017 International
  Conference on Digital Arts, Media and Technology (ICDAMT)}, 2017, pp.
  286--290.

\bibitem{Xu}
H.~Xu, H.~Chen, Y.~Cai, and X.~Liu, ``Underground parking lots layout design
  based on greedy algorithm,'' vol.~25, no.~1, pp. 95--100, 2 2020.

\bibitem{composition}
J.~Togelius, T.~Justinussen, and A.~Hartzen, ``Compositional procedural content
  generation,'' in \emph{Proceedings of The Third Workshop on Procedural
  Content Generation in Games}, ser. PCG'12.\hskip 1em plus 0.5em minus
  0.4em\relax New York, NY, USA: Association for Computing Machinery, 2012, p.
  1–4.

\bibitem{RLpp}
N.~Peng, Y.~Xi, J.~Rao, X.~Ma, and F.~Ren, ``Urban multiple route planning
  model using dynamic programming in reinforcement learning,'' \emph{IEEE
  Transactions on Intelligent Transportation Systems}, vol.~23, no.~7, pp.
  8037--8047, 2022.

\bibitem{WFC}
A.~Sandhu, Z.~Chen, and J.~McCoy, ``Enhancing wave function collapse with
  design-level constraints,'' in \emph{Proceedings of the 14th International
  Conference on the Foundations of Digital Games}, ser. FDG '19.\hskip 1em plus
  0.5em minus 0.4em\relax New York, NY, USA: Association for Computing
  Machinery, 2019.

\bibitem{Akatsu}
N.~Akatsu, M.~Shimizu, and T.~Yonekura, ``A study of the simulation technology
  application to a parking lot layout design,'' in \emph{2014 17th
  International Conference on Network-Based Information Systems}, 2014, pp.
  584--589.

\bibitem{ferns2003metrics}
N.~F. Ferns, ``Metrics for markov decision processes,'' 2003.

\bibitem{lan2022time}
G.~Lan, J.~M. Tomczak, D.~M. Roijers, and A.~Eiben, ``Time efficiency in
  optimization with a bayesian-evolutionary algorithm,'' \emph{Swarm and
  Evolutionary Computation}, vol.~69, p. 100970, 2022.

\bibitem{lan2021learning}
G.~Lan, M.~De~Carlo, F.~van Diggelen, J.~M. Tomczak, D.~M. Roijers, and A.~E.
  Eiben, ``Learning directed locomotion in modular robots with evolvable
  morphologies,'' \emph{Applied Soft Computing}, vol. 111, p. 107688, 2021.

\bibitem{xu2019online}
H.~Xu, G.~Lan, S.~Wu, and Q.~Hao, ``Online intelligent calibration of cameras
  and lidars for autonomous driving systems,'' in \emph{2019 IEEE Intelligent
  Transportation Systems Conference (ITSC)}.\hskip 1em plus 0.5em minus
  0.4em\relax IEEE, 2019, pp. 3913--3920.

\bibitem{lan2019simulated}
G.~Lan, J.~Chen, and A.~E. Eiben, ``Simulated and real-world evolution of
  predator robots,'' in \emph{2019 IEEE Symposium Series on Computational
  Intelligence (SSCI)}.\hskip 1em plus 0.5em minus 0.4em\relax IEEE, 2019, pp.
  1974--1981.

\bibitem{lan2022vision}
G.~Lan, Y.~Wu, F.~Hu, and Q.~Hao, ``Vision-based human pose estimation via deep
  learning: A survey,'' \emph{IEEE Transactions on Human-Machine Systems},
  2022.

\bibitem{sun2022multi}
Z.~Sun, C.~Meng, J.~Cheng, Z.~Zhang, and S.~Chang, ``A multi-scale feature
  pyramid network for detection and instance segmentation of marine ships in
  sar images,'' \emph{Remote Sensing}, vol.~14, no.~24, p. 6312, 2022.

\bibitem{sun2023marine}
Z.~Sun, C.~Meng, T.~Huang, Z.~Zhang, and S.~Chang, ``Marine ship instance
  segmentation by deep neural networks using a global and local attention
  (gala) mechanism,'' \emph{Plos one}, vol.~18, no.~2, p. e0279248, 2023.

\bibitem{lan2022class}
G.~Lan, Z.~Gao, L.~Tong, and T.~Liu, ``Class binarization to neuroevolution for
  multiclass classification,'' \emph{Neural Computing and Applications},
  vol.~34, no.~22, pp. 19\,845--19\,862, 2022.

\bibitem{gao2021neat}
Z.~Gao and G.~Lan, ``A neat-based multiclass classification method with class
  binarization,'' in \emph{Proceedings of the genetic and evolutionary
  computation conference companion}, 2021, pp. 277--278.

\bibitem{liu2022towards}
T.~Liu, G.~Lan, K.~A. Feenstra, Z.~Huang, and J.~Heringa, ``Towards a knowledge
  graph for pre-/probiotics and microbiota--gut--brain axis diseases,''
  \emph{Scientific Reports}, vol.~12, no.~1, p. 18977, 2022.

\bibitem{lan2022semantic}
G.~Lan, T.~Liu, X.~Wang, X.~Pan, and Z.~Huang, ``A semantic web technology
  index,'' \emph{Scientific reports}, vol.~12, no.~1, p. 3672, 2022.

\bibitem{lan2019evolving}
G.~Lan, L.~De~Vries, and S.~Wang, ``Evolving efficient deep neural networks for
  real-time object recognition,'' in \emph{2019 IEEE Symposium Series on
  Computational Intelligence (SSCI)}.\hskip 1em plus 0.5em minus 0.4em\relax
  IEEE, 2019, pp. 2571--2578.

\bibitem{lan2018real}
G.~Lan, J.~Benito-Picazo, D.~M. Roijers, E.~Dom{\'\i}nguez, and A.~Eiben,
  ``Real-time robot vision on low-performance computing hardware,'' in
  \emph{2018 15th international conference on control, automation, robotics and
  vision (ICARCV)}.\hskip 1em plus 0.5em minus 0.4em\relax IEEE, 2018, pp.
  1959--1965.

\end{thebibliography}

\end{document}